# Anti-Money Laundering Machine Learning Pipelines; A Technical Analysis on Identifying High-risk Bank Clients with Supervised Learning

Khashayar Namdar[a,d,e]*, Pin-Chien Wang[b], Tushar Raju[b], Steven Zheng[b], Fiona Li[b], Safwat Tahmin Khan[c]

[a]Institute of Medical Science, University of Toronto, Toronto, ON, Canada; [b]Rotman School of Management, University of Toronto, Toronto, ON, Canada; [c]Institute of Biomedical Engineering, University of Toronto, Toronto, ON, Canada; [d]Vector Institute, Toronto, ON, Canada; [e]NVIDIA Deep Learning Institute, Austin, TX, United States

*corresponding author: Khashayar Namdar, ernest.namdar@utoronto.ca

# Anti-Money Laundering Machine Learning Pipelines; A Technical Analysis on Identifying High-risk Bank Clients with Supervised Learning

Anti-money laundering (AML) actions and measurements are among the priorities of financial institutions, for which machine learning (ML) has shown to have a high potential. In this paper, we propose a comprehensive and systematic approach for developing ML pipelines to identify high-risk bank clients in a dataset curated for Task 1 of the University of Toronto 2023-2024 Institute for Management and Innovation (IMI) Big Data and Artificial Intelligence Competition. The dataset included 195,789 customer IDs, and we employed a 16-step design and statistical analysis to ensure the final pipeline was robust. We also framed the data in a SQLite database, developed SQL-based feature engineering algorithms, connected our pre-trained model to the database, and made it inference-ready, and provided explainable artificial intelligence (XAI) modules to derive feature importance. Our pipeline achieved a mean area under the receiver operating characteristic curve (AUROC) of 0.961 with a standard deviation (SD) of 0.005. The proposed pipeline achieved second place in the competition.

**Introduction**

In the contemporary financial landscape, money laundering represents a formidable challenge, compelling both financial institutions and regulatory bodies to seek innovative solutions. The integration of machine learning (ML) into anti-money laundering (AML) efforts has emerged as a promising avenue to enhance the detection and prevention of illicit financial activities. This paper investigates the technical considerations in employing supervised learning techniques to accurately identify high-risk bank clients, a critical component in the battle against money laundering.

The utilization of ML for detecting money laundering transactions has shown significant promise. Jullum et al. developed an ML model that outperforms traditional systems by prioritizing transactions for manual investigation, using historic data from Norway's largest bank, DNB [1]. Alarab, Prakoonwit, and Nacer further demonstrated the efficacy of supervised learning methods in identifying licit and illicit transactions within the Bitcoin blockchain, showcasing an ensemble learning method that surpassed classical approaches [2]. Additionally, Sintayehu and Seid highlighted the superior performance of Random Forest (RF) algorithm in identifying money laundering activities within a dataset from the Commercial Bank of Ethiopia, achieving an accuracy of 99.1% [3].

This body of work underscores the pivotal role of ML, particularly supervised learning techniques, in refining the detection of high-risk clients and transactions within the banking sector. The pipelines are developed and validated based on the dataset curated by Scotiabank for Task 1 of the University of Toronto 2023-2024 Institute for Management and Innovation (IMI) Big Data and Artificial Intelligence Competition. Nevertheless, the methodology can be applied to any supervised AML high-risk client identification scenario. By leveraging the computational power and pattern recognition capabilities of ML algorithms, financial institutions can significantly enhance their AML strategies, leading to more effective and efficient prevention of money laundering activities.

**Materials and Methods**

*Dataset*

The dataset included 4 components: Know Your Customer (kyc.csv), cash transactions (cash_trxns.csv), email transfers (emt_trxns.csv), and wire transfers (wire_trxns.csv). Task 1 of the University of Toronto 2023-2024 Institute for Management and Innovation (IMI) Big Data and Artificial Intelligence Competition involved supervised learning for identifying high-risk customers. The ground truth binary labels were provided in KYC through the "label" columns for 195,789 customers. KYC also included name, gender, occupation, age, tenur, and customer id (cust_id). Table 1 summarizes the features and their corresponding data types.

Table 1. Features and data types

| File | Column | Data Type | Notes |
|---|---|---|---|
| kyc | Name | str | There are repeated names<br><br>145,393 unique values |
| kyc | Gender | str | 'female' 'male' 'other' |
| kyc | Occupation | str | 250 unique values |
| kyc | Age | int | 75 unique integer values (min=18, max=92) |
| kyc | Tenur | int | refers to the length of time a customer has had a relationship with the financial institution or service provider<br><br>49 unique integer values (min=0, max=49) |
| kyc | cust_id | str | cust_ids are all unique (number of rows in the sheet = 195,789)<br><br>195,789 unique values |
| kyc | label | int (binary) | [0, 1] 1 refers to high risk customers |
| cash_trxns | cust_id | str | Same as kyc cust_id |
| cash_trxns | amount | int | |
| cash_trxns | type | str | Binary: ["deposit", "withdrawal"] |
| cash_trxns | trxn_id | str | Transaction id |
| emt_trxns | id sender | str | |
| emt_trxns | id receiver | str | |
| emt_trxns | name sender | str | |
| emt_trxns | name receiver | str | |
| emt_trxns | emt message | str | |
| emt_trxns | emt value | float | |
| emt_trxns | trxn_id | str | Transaction id |
| wire_trxns | id sender | str | |
| wire_trxns | id receiver | str | |
| wire_trxns | name sender | str | |
| wire_trxns | name receiver | str | |
| wire_trxns | wire value | float | |
| wire_trxns | country sender | str | |
| wire_trxns | country receiver | str | |
| wire_trxns | trxn_id | str | Transaction id |

*Base Pipeline*

In the process of selecting an appropriate base pipeline for our computational model, the decision tree algorithm was chosen owing to its fundamental simplicity and the distinct advantage of reduced training duration. Decision trees (DT) offer an intuitive structure that is facile both in comprehension and interpretation, rendering them a compelling option for preliminary modeling endeavors. The intrinsic simplicity of DTs obviates the need for substantial computational resources, thereby diminishing the training period necessary to formulate an efficacious model. Consequently, DTs emerge as a prime candidate for the foundational layer in the construction of predictive models, especially in contexts demanding swift development and expedited deployment. This preference is underpinned by literature that attests to the computational efficiency and ease of use of decision trees, notably in the manipulation of large datasets [4]. We evaluate the performance of the models based on different metrics such as accuracy, precision, recall, and tools such as the confusion matrix. However, a single and comprehensive evaluation metric streamlines judging the pipelines and conducting statistical analysis. Given the fact that the labels are binary and represent risk, which can be considered a binarized continuous variable, we chose the area under the receiver operating characteristic curve (AUROC) as the evaluation metric [5]. For the final model, we will provide all the common evaluation metrics in ML. For the base pipeline, we limited the models to KYC.

*Measuring Randomness*

Measuring randomness is a crucial aspect in various data science processes. In supervised AML, major sources of randomness are data splitting and model initialization [6]. Measuring randomness ensures that models are not biased towards particular patterns or distributions within the data. In data splitting, randomness is measured to achieve representative train-test splits, ensuring that both subsets reflect the overall dataset's characteristics. Similarly, model initialization often relies on random parameter assignments to prevent the model from converging to suboptimal solutions. Quantifying randomness is integral to the integrity of A/B testing, allowing for a fair and accurate comparison of different versions or strategies. At this stage, we use the Monte Carlo method to measure the performance of the pipeline over different data splits and with varied random states. This enables statistical tests to compare different designs.

*Running Time*

Measuring the running time of ML pipelines is particularly crucial when dealing with big data, as it directly impacts scalability and efficiency. Long running times can significantly delay insights and hinder the iterative process of model refinement and deployment. Therefore, understanding and optimizing the computational performance of ML pipelines ensures that they can handle large volumes of data effectively, facilitating timely decision-making and enabling more complex analyses within reasonable time frames. For a major financial institute, a marginal accuracy improvement at the expense of significant inference delays is not acceptable. We timed all the candidate models and used running time as a factor for choosing the final model.

*Switching to Random Forests*

Random Forest (RF) models are shown to be powerful for complicated tasks such as radiomics classification [7]. Switching from a decision tree to a random forest is beneficial for big tabular data because it aggregates multiple trees to reduce overfitting and improve prediction accuracy by capturing more complex patterns and interactions within the data. We trained RF models and compared them against the DT models.

*Hyperparameter Tuning*

Hyperparameters (e.g., number of trees in an RF) are non-learnable parameters of ML models that directly affect their design and performance. For a simple algorithm such as DT, hyperparameter tuning may not be a major concern. However, the vast majority of ML algorithms, including RF models, will be suboptimal without hyperparameter tuning. We implemented a Monte-Carlo-based approach for grid search, and kept the test cohorts unseen.

*XGBoost*

Extreme Gradient Boosting (XGBoost) has emerged as a state-of-the-art (SOTA) ML algorithm, particularly renowned for its performance in handling structured or tabular data. Unlike random forests, which rely on averaging numerous decision tree outputs to enhance prediction accuracy, XGBoost employs a more sophisticated approach called gradient boosting. This method optimizes for a specific loss function iteratively, allowing it to efficiently handle both bias and variance, leading to superior performance. Moreover, XGBoost incorporates built-in features for regularizing model complexity, which further mitigates the risk of overfitting. Its capacity to deal with missing data, alongside its scalability and parallel processing capabilities, makes XGBoost a more powerful and versatile choice than RF, especially in competitive ML tasks. We utilized XGboost and compared it against RF.

*Data Splitting*

Data splitting is a fundamental technique in ML that divides the dataset into separate sets, typically for training, validation, and testing. This process is critical for evaluating the performance of a model in an unbiased manner. Monte Carlo cross-validation randomly splits the data into training and testing sets multiple times and averages the results to assess model performance, providing flexibility and randomness. In contrast, K-fold cross-validation systematically divides the data into K equally sized folds, using each fold as a testing set while the remaining serve as the training set, thereby ensuring that every data point is used for both training and testing. While Monte Carlo allows for multiple iterations with random splits, potentially leading to different results each time, K-fold offers a more structured approach, ensuring each sample is tested exactly once. If the dataset is very large, the two methods are expected to result in the same performance level. Nevertheless, we compare the two techniques for developing our pipeline.

*Encoding*

Label encoding and one-hot encoding are two techniques for converting categorical data into numerical format, essential for machine learning models. Label encoding assigns a unique integer to each category, transforming the categorical variable into a single numerical column. This method is simple and efficient, but can imply an ordinal relationship between categories where none exists. On the other hand, one-hot encoding creates a separate binary column for each category, representing the presence or absence of a category with a 1 or 0, respectively. This approach avoids implying any false ordinality but can significantly increase the dataset's dimensionality. We compared the two techniques when transforming gender and occupation from KYC.

*State-of-the-Art Models*

ML is an established field, and there are numerous algorithms that can be used for the task of tabular data classification. We implemented and compared different SOTA algorithms to ensure a reliable and accurate pipeline was proposed.

XGBoost, CatBoost, LightGBM, and TabNet are prominent ML algorithms that excel in processing tabular data. XGBoost is celebrated for its scalability and efficiency in handling sparse data, making it a staple in data science competitions [8]. CatBoost is distinguished for its ability to handle categorical data directly, reducing preprocessing time and potentially increasing model accuracy [9]. LightGBM stands out for its speed and efficiency, particularly in large datasets, due to its gradient-based one-sided sampling [10]. TabNet, utilizing neural networks (NN), offers a unique approach by leveraging attention mechanisms for tabular data, providing interpretability similar to decision trees while maintaining the flexibility of neural networks [11].

AutoML represents a paradigm shift in ML, aiming to automate the process of applying ML to real-world problems [12]. AutoGluon, specifically, stands as a powerful AutoML tool designed to achieve high accuracy with minimal user input [13]. It simplifies model selection, hyperparameter tuning, and ensemble creation, making state-of-the-art ML models accessible to non-experts. AutoGluon's versatility and ease of use democratize the application of ML, bridging the gap between complex ML tasks and practical applications. In order to ensure the optimal algorithm was included, we used AutoGluon in addition to the individual models that were tested.

*Imbalanced Classes*

Imbalanced data significantly impacts ML models, often resulting in models that are biased towards the majority class and perform poorly on the minority class. Strategies such as undersampling the majority class can reduce the data imbalance but may lead to loss of valuable information. Oversampling the minority class addresses this by replicating the minority class instances, yet it may increase the risk of overfitting due to the exact replication of instances. Synthetic Minority Over-sampling Technique

(SMOTE) offers a refined approach by generating synthetic samples based on the feature space similarities between the minority class instances, potentially improving the generalization ability of models on imbalanced datasets [14]. By creating these synthetic samples, SMOTE mitigates the risk of overfitting associated with simple oversampling and preserves information better than undersampling, making it a more balanced and effective approach for training machine learning models on imbalanced datasets. We instigate all these approaches to achieve a reliable pipeline.

*Dataset Size Sensitivity*

Dataset size sensitivity refers to how the performance of ML models is influenced by the volume of data available for training. Models can exhibit varying degrees of accuracy, generalizability, and overfitting risk depending on the dataset's size, indicating the critical role that the quantity of data plays in achieving optimal model performance [15]. For small datasets, dataset size sensitivity verifies the power calculations, and with big data, it can help achieve faster pipelines by excluding unneeded examples from the training set. We conducted a comprehensive dataset size sensitivity, augmented with randomness measurement.

*Mega Test*

In some ML scenarios (e.g., following dataset size sensitivity analysis and undersampling the larger class in an imbalanced dataset) a proportion of the training set is discarded. Depending on the evaluation method, the discarded examples can be appended to the test set. Using AUROC as the evaluation metric enabled conducting a mega test using the combination of the initial test set and the unused training examples.

*Switching to Databases*

Real-world AML piplelines have to interact with existing databases of financial institutions. ML algorithms for AML that use spreadsheets as input are preliminary models with limited capacity for external evaluation and dataset evolution. To manage the existing tabular data, Structured Query Language (SQL) is an appropriate choice. We developed Python scripts and converted the spreadsheets into an SQLite database, as illustrated in Fig. 1.

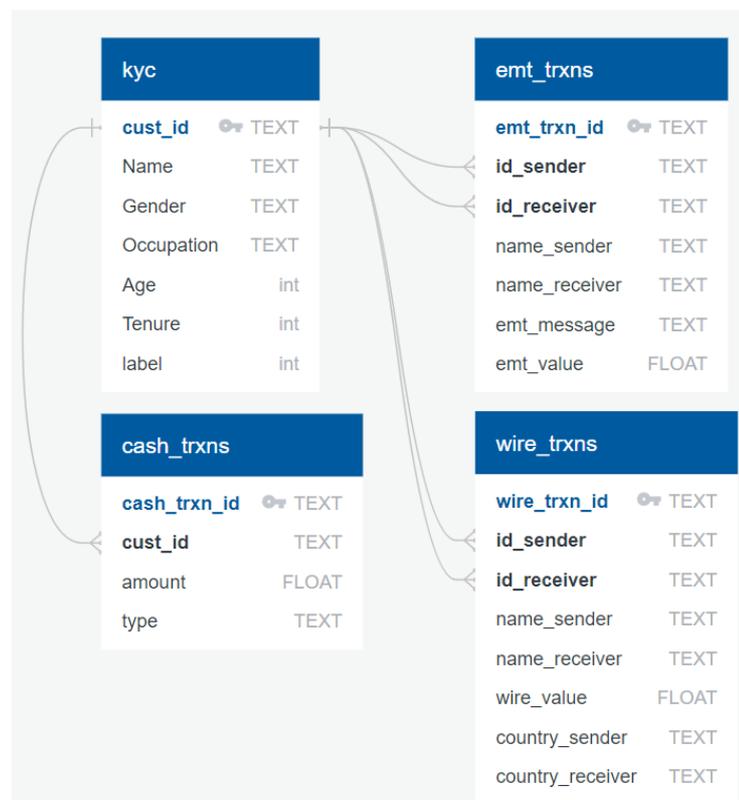

Figure 1. Gender distribution by label

*Feature Engineering*

Feature engineering for tabular data classification involves the process of selecting, modifying, or creating new features from raw data to improve the performance of ML models. This crucial step can include techniques such as normalization, fusion, one-hot encoding, and dealing with missing values to enhance model accuracy and efficiency. Although we investigated two techniques in the encoding section, our feature engineering process was not comprehensive. KYC is a portion of the available data that is straightforward to be used for the supervised classification task. Nonetheless, the available transaction data can improve the models. SQL streamlines feature engineering and thus, we implemented three different versions of feature engineering.

In Version 1 of the feature engineering process, the focus was on constructing a comprehensive feature set encapsulating customer transaction behaviors. This set includes metrics on the number and total amount of wire transactions (both sent and received, including a distinction between domestic and international transfers), EMT transactions (both sent and received), as well as cash deposits and withdrawals. A noteworthy aspect of this feature engineering effort is the handling of non-occurrences: for any type of transaction that a customer has not performed, corresponding fields are populated with zeros. This approach ensures data consistency and prevents issues that might arise from null values, thereby facilitating a more straightforward analysis and modeling process.

Version 2 of the feature engineering process introduces significant enhancements over Version 1, notably including aggregate metrics for international, wire, EMT, and cash transactions, as well as balance calculations for each transaction type. These additions provide a more nuanced view of customer transaction behaviors by capturing overall transaction volumes and net transaction balances, offering insights into financial habits and risk profiles. While both versions ensure data consistency by handling non-occurrences with zeros, Version 2 streamlines the data further by integrating these new metrics directly into the dataset, thereby enriching the analytical depth and facilitating a more comprehensive analysis without additional data manipulation.

Version 3 enhances the feature engineering process established in Version 2 by integrating a geographical layer into the analysis, focusing on the countries involved in wire transactions. It retains the core metrics of Version 2 while introducing country-specific transaction counts for both sent and received wire transactions, using a predefined list of countries. For transactions involving countries not on this list, a 'NAN' category is used. This development allows for a detailed exploration of international transaction patterns, providing insights into customer preferences and behaviors across different geographical regions. By tracking transactions to and from specified countries, this version offers a richer, more nuanced dataset capable of supporting in-depth financial analyses, such as identifying trends in international dealings, assessing risk exposure, and detecting potential fraud, thereby significantly enhancing the dataset's analytical depth and applicability for predictive modeling.

***Back to Single Model***

Once a robust and reliable pipeline is developed and its randomness is measured, a single model can be trained to be deployed and used for inference. An alternative approach is training multiple models and employing an ensemble method. However, we chose a single model because the performance was near ideal, the dataset size was enormous, and inference time was a decisive factor. Switching to a single model enabled calculating multiple evaluation metrics and conducting explainable artificial intelligence (XAI) analyses to derive feature importance for the model.

*Continuous Machine Learning*

Continuous Machine Learning (CML) is a pivotal component in maintaining the relevance and accuracy of predictive models in dynamic environments. By automating the training, evaluation, and deployment of models with new data, CML ensures that predictions remain accurate over time, even as underlying patterns within the data evolve. This approach enables organizations to rapidly adapt to changes, such as emerging trends, seasonal variations, or unforeseen events, thereby securing a competitive edge and enhancing decision-making processes. Furthermore, CML fosters a culture of continuous improvement and innovation by facilitating the iterative refinement of models, encouraging experimentation, and reducing the time and resources required to deploy improvements. In fields where timely and precise predictions are critical, such as fraud detection, market forecasting, and personalized recommendations, the importance of CML cannot be overstated, as it directly contributes to operational efficiency, customer satisfaction, and overall success. We developed a CML module to enable updating the model for inference. This can be triggered based on a timestamp or number of changes in the database.

*Graphical User Interface*

The incorporation of a Graphical User Interface (GUI) in the high-risk customer identification and Anti-Money Laundering (AML) pipeline is of paramount importance, serving as a critical interface that significantly enhances the usability, efficiency, and interpretability of complex analytical processes. By providing a visual and interactive platform, a GUI facilitates a more intuitive interaction with the underlying data and models, enabling users to seamlessly navigate through vast datasets, apply filters, and visualize results in a comprehensible manner. This enhanced accessibility and ease of use are crucial for stakeholders who may not have deep technical expertise, allowing them to effectively participate in and contribute to the detection and investigation processes. Moreover, the integration of a GUI supports real-time feedback and adjustments, fostering a dynamic environment where insights and anomalies can be promptly identified and acted upon. In the context of identifying high-risk customers and combating money laundering, the GUI acts as a bridge between sophisticated data-driven algorithms and practical, actionable intelligence, thereby playing a vital role in strengthening the overall effectiveness and responsiveness of the AML pipeline. We implemented a web-based interface to deliver the pipeline to the end-users in the financial institutes.

**Results and Discussion**

In our study, we evaluated ML pipelines designed to identify high-risk bank clients for anti-money laundering (AML) efforts. Utilizing the dataset of 195,789 customer IDs from the University of Toronto's IMI Big Data and Artificial Intelligence Competition, our approach incorporated SQL-based feature engineering and XAI modules. All codes are available through our GitHub repository (https://github.com/knamdar/IMI_BigData_AI_Hub_Competition_2023_2024).

In the data exploration phase, we investigated gender distribution (Fig 2), occupations with the highest percentage of risky customers (Fig 3), Age distribution (Fig 4), Tenur distribution (Fig 5), and compared the class sizes (Fig 6) using the label column in KYC. The data exploration did not show any strong differentiator signal, but the first bins of age and tenur histograms made them candidates for strong predictors. Also, the occupations had different proportions of the two classes, and thus occupation is another important predictor in KYC. Repeated names did not show to have any significant relationship with the labels.

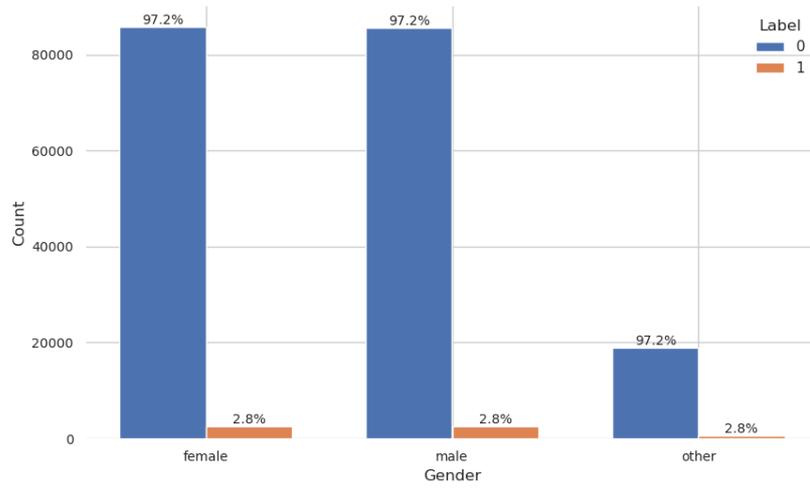

Figure 2. Gender distribution by label

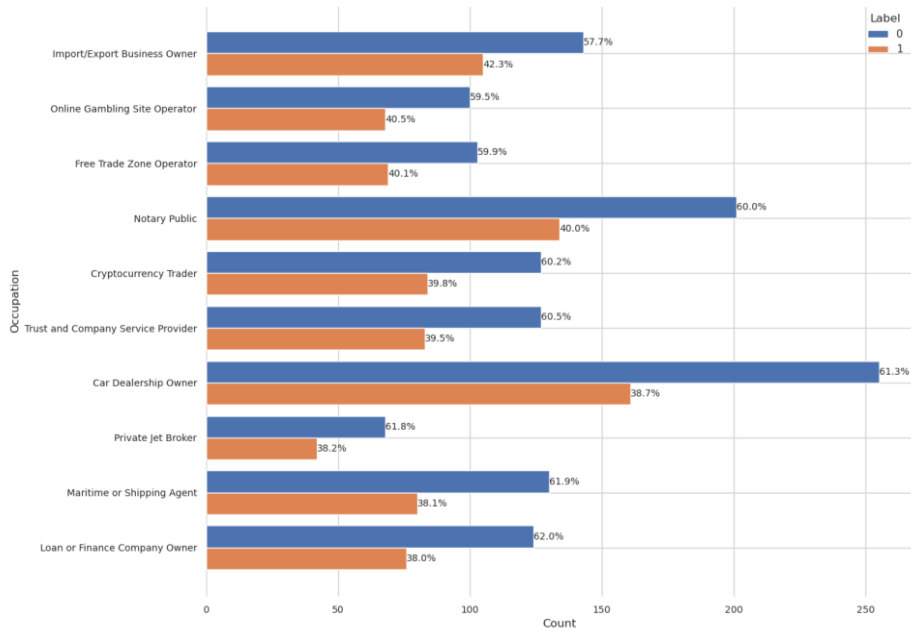

Figure 3. Top 10 occupations with the highest percentage of risky customers

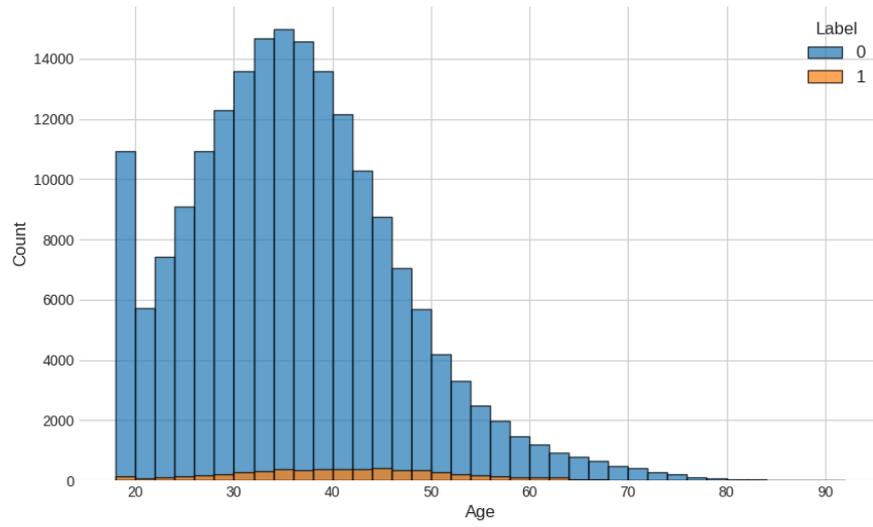

Figure 4. Age distribution by label

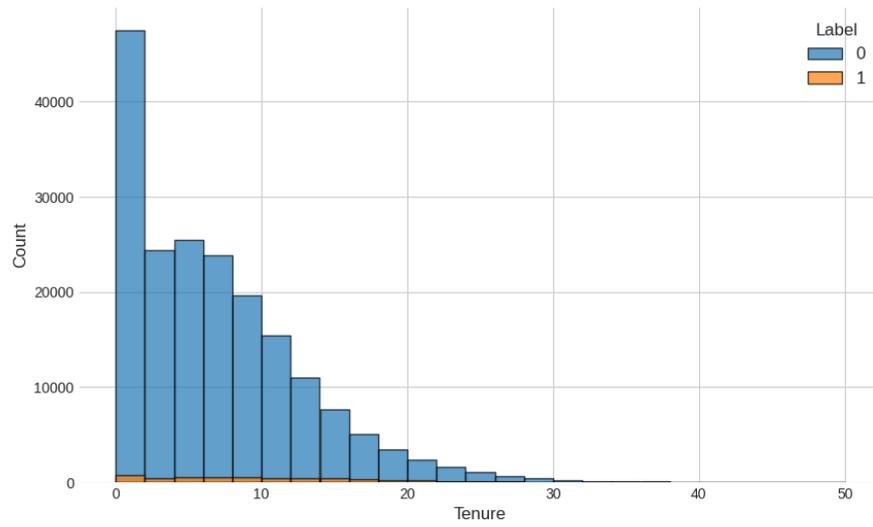

Figure 5. Tenur distribution by label

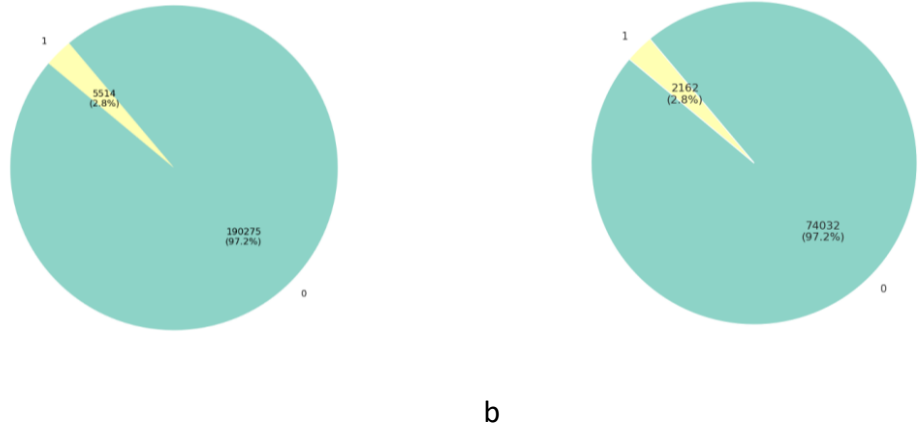

a    b

Figure 6. Class size visualization a) whole dataset b) customers with non-unique names

We employed a modular approach for designing the pipeline, which is illustrated in Fig. 7. The base pipeline was based on a DT model, a test-to-train ratio of 25%/75% with a singe random stratified data split, and resulted in an AUROC of 0.592. Although we achieved an accuracy of 0.958, it was not reliable because data exploration showed the ratio of the largest class to the dataset size was 0.972. Therefore, we implemented a random undersampling method and applied it upfront to the whole dataset to lower the size of the low-risk class and perfectly balance the dataset. On the balanced dataset, the AUROC improved to AUC 0.678. The corresponding confusion matrices are shown in Fig. 8. This experiment demonstrated the KYC features could potentially differentiate the data. Also, the effect of the imbalanced classes was found to be prominent and needs to be taken into account.

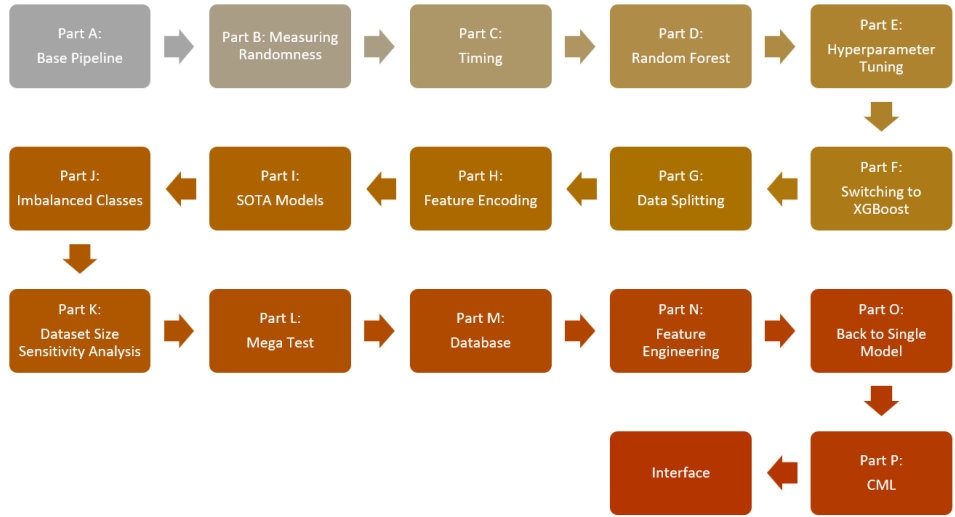

Figure 7. Overview of the design procedure

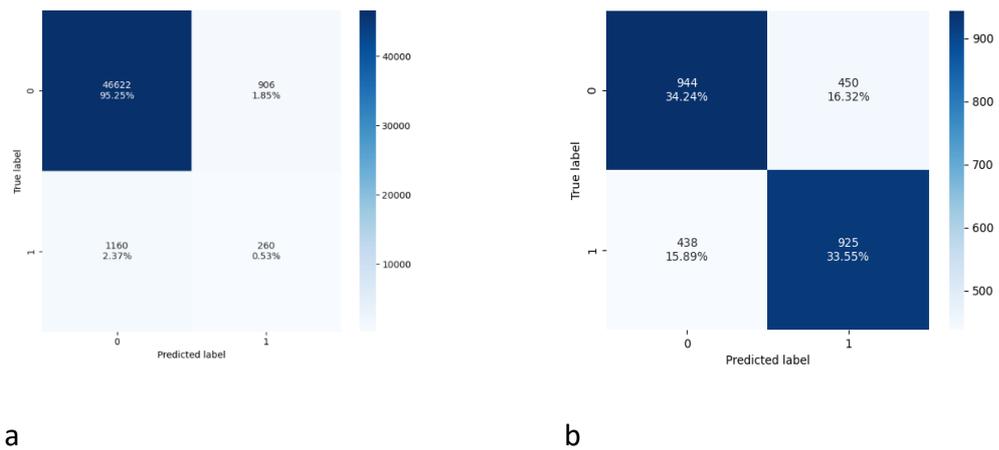

Figure 8. Confusion Matrices of the base models: a) whole dataset b) balanced dataset

Using the same settings as before, we implemented a Monte Carlo based method for repeating the experiment 30 times. In each repeat, random state of the models and the data split were changed. Fig. 9 compares the AUROC performance of the DT models on the whole dataset and the balanced version where mean AUROCs were 0.598 and 0.699, for whole and balanced datasets, respectively. For the statistical analysis, we used Student's t test. Based on the results, we limited the experiments to the balanced dataset, until the stage where we tackled data imbalanceness. Timing the operations showed that DT training and evaluation in both cases were almost instant.

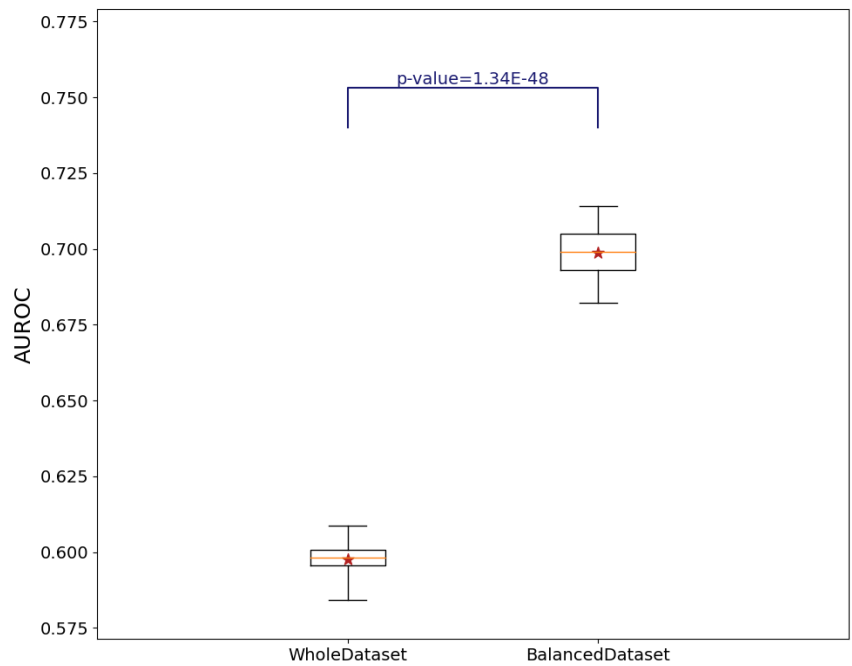

Figure 9. Comparing the performance of the DT models with and without balancing the classes

DT did not have the capacity to accurately differentiate the data and resulted in underfitting. Thus, we trained RF models and compared them against DTs (Fig. 10). RF improved the mean AUROC to 0.760, and thus was chosen as the baseline algorithm for the next experiments.

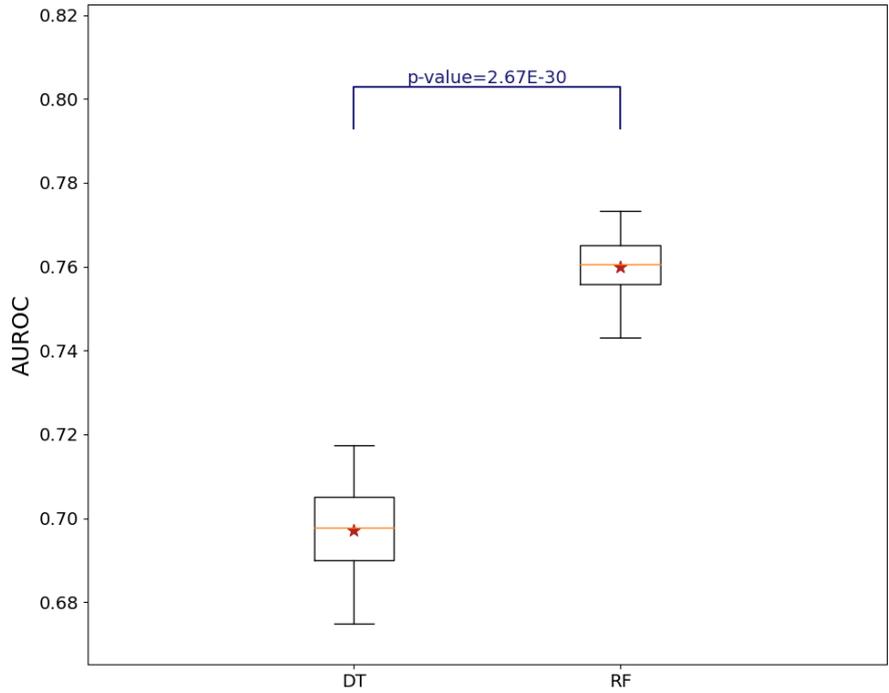

Figure 10. Comparing the performance of DT and RF

When training the RF, we used the default settings of the scikit-learn Version 1.3.2. However, hyperparameter tuning is a must in ML and thus we implemented a Monte Carlo based method for conducting it. In each experiment of the 30 experiments with 25%/75% test-to-development ratio, we conducted an inner loop and performed 30 validations with 25%/75% validation-to-train ratio on the development sets. Table 2 shows the grid space.

Table 2. Grid space of the RF models

| Hyperparameter | Candidate Values |
|---|---|
| n_estimators | 50, 100, 200 |
| max_features | 'auto', 'sqrt' |
| max_depth | None, 5, 10 |

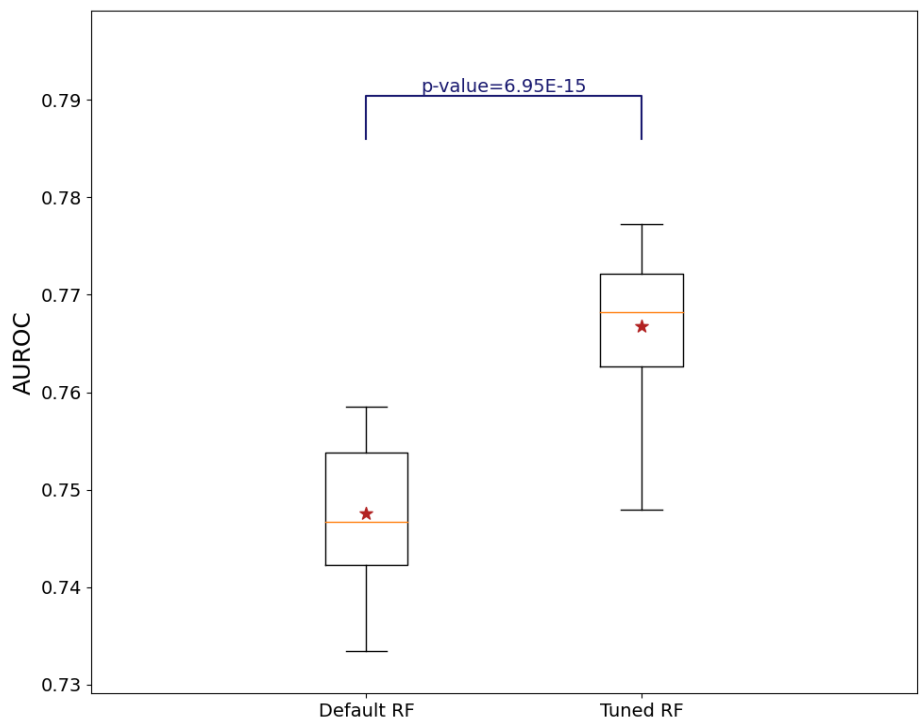

Figure 10. Comparing the performance of RF with and without hyperparameter tuning

Hyperparameter tuning improved the mean AUROC of the models from 0.748 to 0.767 which was statistically significant. However, the execution time was increased from 21 seconds to over 2 hours. This is because 30 training-inference experiments were increased to 16200 (30 tests x 30 validations x 18 hyperparameter sets). Note that we reported a mean AUROC of 0.760 when comparing RF with DT, but at this stage, it is 0.748. The discrepancy is caused by the randomness of the pipeline.

Switching from RF to XGB significantly improved the mean AUROC from 0.771 to 0.862 (Fig. 11), while decreasing the execution time from 123 minutes to 62 minutes. Nevertheless, this should not be interpreted as the simplicity of XGBoost. As shown in Fig. 12, XGBoost features a more advanced implementation and utilizes all the available resources for parallel processing. If needed, parallel processing can be implemented for RF. The results updated our baseline model to XGBoost.

Table 3. Grid space of the XGBoost models

| Hyperparameter | Candidate Values |
| --- | --- |
| n_estimators | 50, 100, 200 |
| learning_rate | 0.01, 0.1, 0.2 |
| max_depth | 5, 10 |

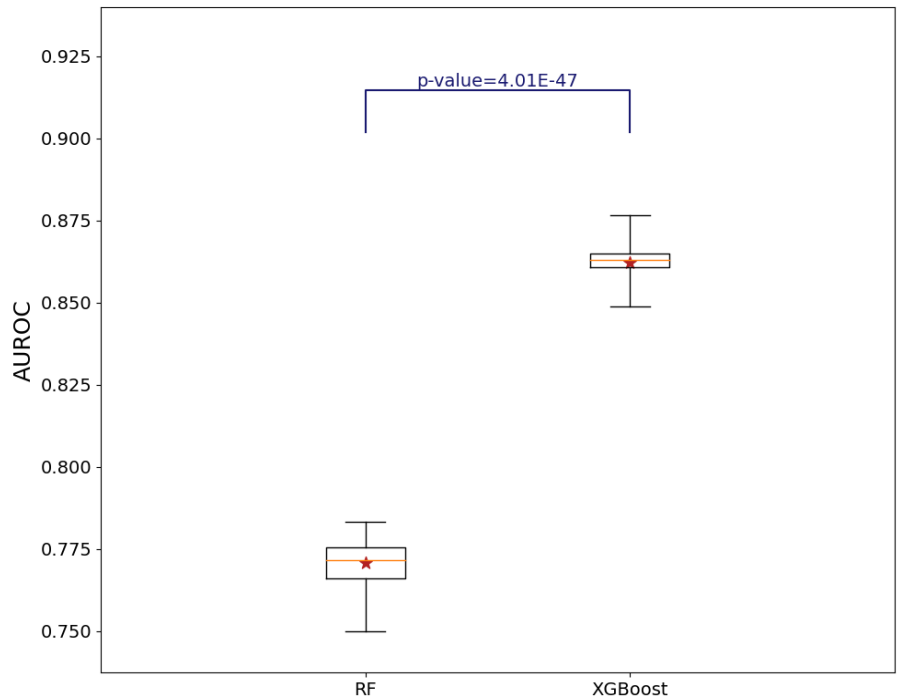

Figure 11. Comparing the performance of RF with XGBoost

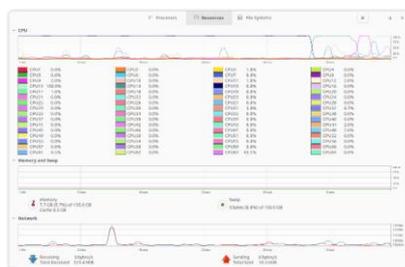
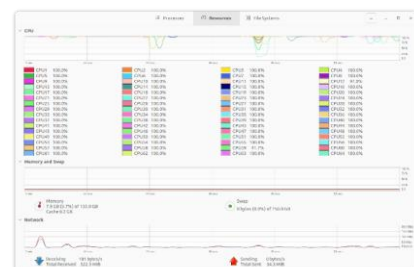

a                                    b

Figure 12. Comparing the resource utilization of a) RF with b) XGBoost

Given that the run times were relatively high, we explored KFold cross-validation to save time while conducting the experiments (Fig. 13). We implemented a nested 10-fold cross-validation (10 folds for test/development and 10 folds for validation/train). Although the results were not statistically significantly different (mean AUROC of 0.860 versus 0.863 for Monte Carlo and KFold, respectively), the run time was improved from 61 minutes to 6 minutes. Thus, we switched the design of the pipeline to nested 10-Fold cross-validation.

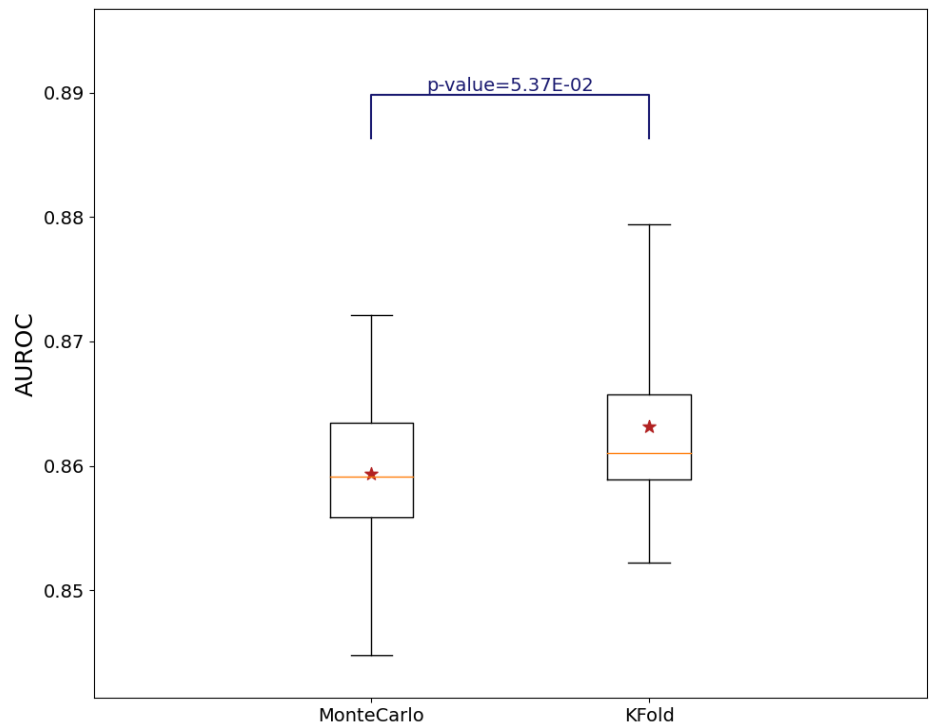

Figure 13. Comparing the Monte Carlo and KFold data splitting techniques

One-hot encoding significantly outperformed label encoding (mean AUROC 0.858 vs 0.869), but it also increased the execution time from 6 minutes and 40 seconds to 10 minutes and 54 seconds. Given the size of the dataset and the number of features (large n, small p), it was expected that one-hot encoding help the results. Since the final model will have one time training, we chose to use one-hot encoding.

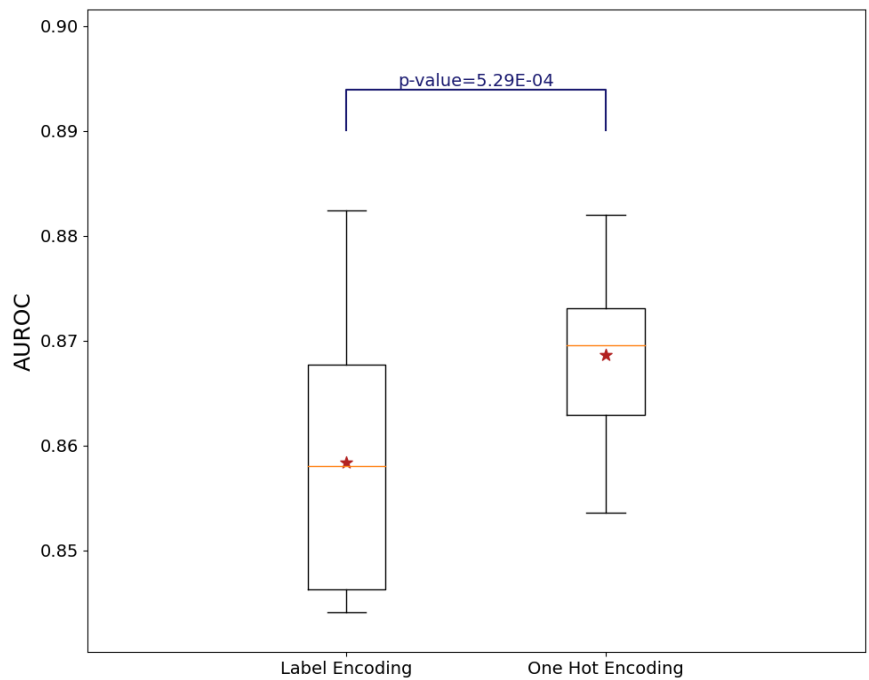

Figure 14. Comparing the Label Encoding with One Hot Encoding for gender and occupation

In the next step, we compared different SOTA models (Fig. 15). First, CatBoost significantly outperformed XGBoost (mean AUROC 0.858 vs 0.864 and run time 11 min vs 27 min). Then LightGBM significantly outperformed CatBoost while demonstrating an excellent execution time (mean AUROC 0.862 vs 0.871 and run time 27 min vs 4 min). The run time of TabNet did not allow hyperparameter tuning. TabNet with a max_epochs of 100, learning rate of 5e-2, batch size of 256 and virtual batch size of 128, resulted in a mean AUROC of 0.875 in 17 minutes, which was not a significant improvement. Given its run time, TabNet was not selected as the baseline model. Lastly, AutoGluon achieved a mean AUROC of 0.872 in 504 minutes. Although the run time is extremely high, it is a one time run. However, LightGBM has a similar performance level and thus was chosen as the target model.

Table 4. Grid space of the CatBoost models

| Hyperparameter | Candidate Values |
|---|---|
| iterations | 50, 100, 200 |
| learning_rate | 0.01, 0.1, 0.2 |
| depth | 5, 10 |

Table 5. Grid space of the LightGBM models

| Hyperparameter | Candidate Values |
|---|---|
| n_estimators | 50, 100, 200 |
| learning_rate | 0.01, 0.1, 0.2 |
| max_depth | 5, -1 |

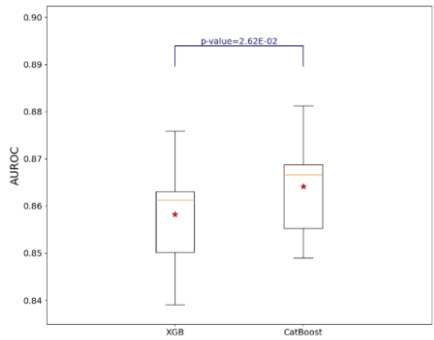

a

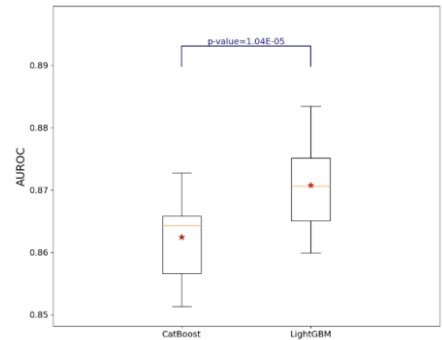

b

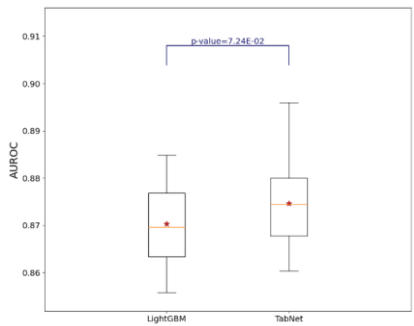

c

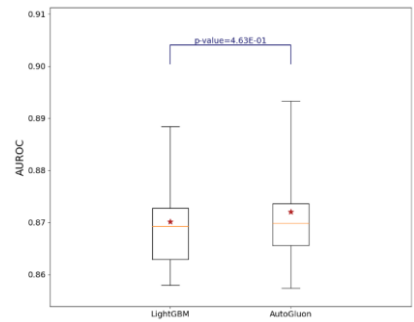

d

Figure 15. Comparing SOTA models a) XGBoost vs CatBoost b) CatBoost vs LightGBM c) LightGBM vs TabNet d) LightGBM vs AutoGluon

We instigated different approaches to tackle the imbalanced classes and none of the methods appeared to be significantly different (Fig. 16). First, we compared undersampling the larger class in the whole dataset upfront (as before) versus training a tuned model on the whole dataset (boosting_type: gbdt, objective: binary, metric: auc, is_unbalance: True). As it was expected, this increased the run time from 4 minutes to 19 minutes. The results showed the effectiveness of the applied tunings, and thus we kept them for the rest of the experiments.

Then we compared undersampling the larger class and oversampling the smaller class, both applied to the development set thanks to the tuned hyperparameters (boosting_type: gbdt, objective: binary, metric: auc, is_unbalance: True). This eliminated the bias of balancing the whole dataset upfront (manipulating the test set indirectly). The execution time increased from 4 to 32 minutes. Finally SMOTE increased the run time from 4 to 44. At the end of this stage, the pipeline was updated to undersampling the majority class on the development set, and introducing new settings for the LightGBM hyperparameters.

a

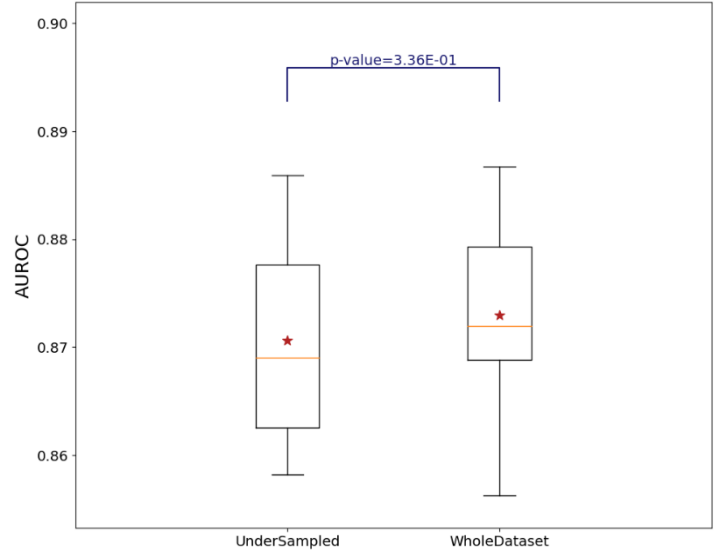

b

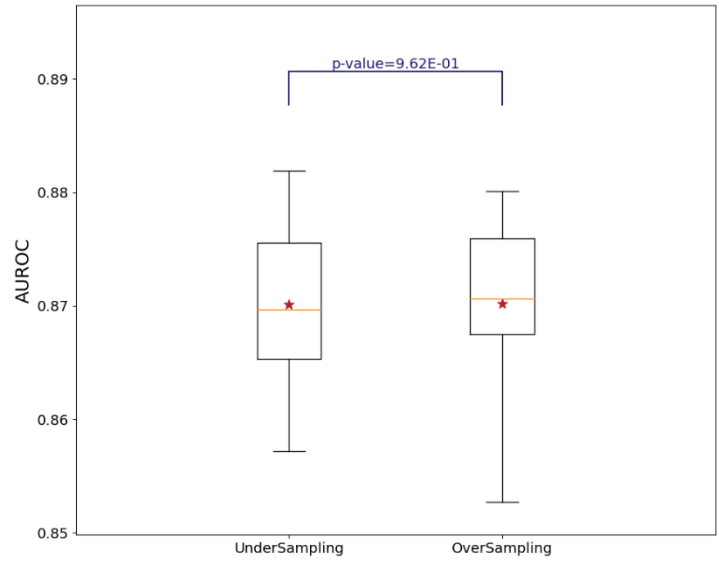

c

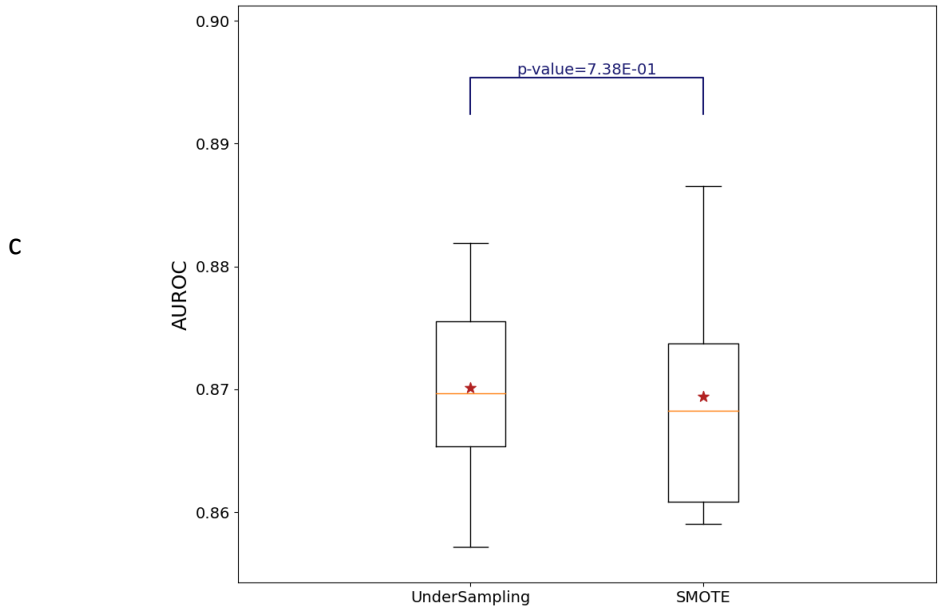

Figure 16. Tackling imbalanced classes

The dataset size sensitivity experiments showed that 5000 examples in the training set would suffice to achieve an optimal model (Fig. 17). This can lower the run time of training the model and help plan the regular model fine-tunings after deployment.

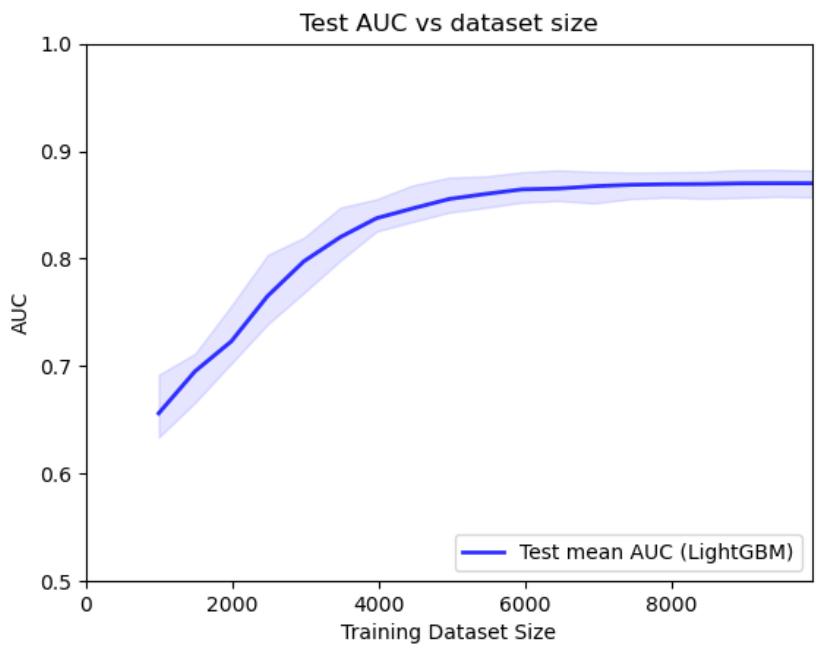

Figure 17. Dataset size sensitivity analysis

Appending the discarded examples in undersampling to the test cohort and conducting a mega test (Fig. 18) did not result in significantly different mean AUROCs or run times (AUROC 0.870 and run time 4 minutes in both cases). This ensured reliability of the models.

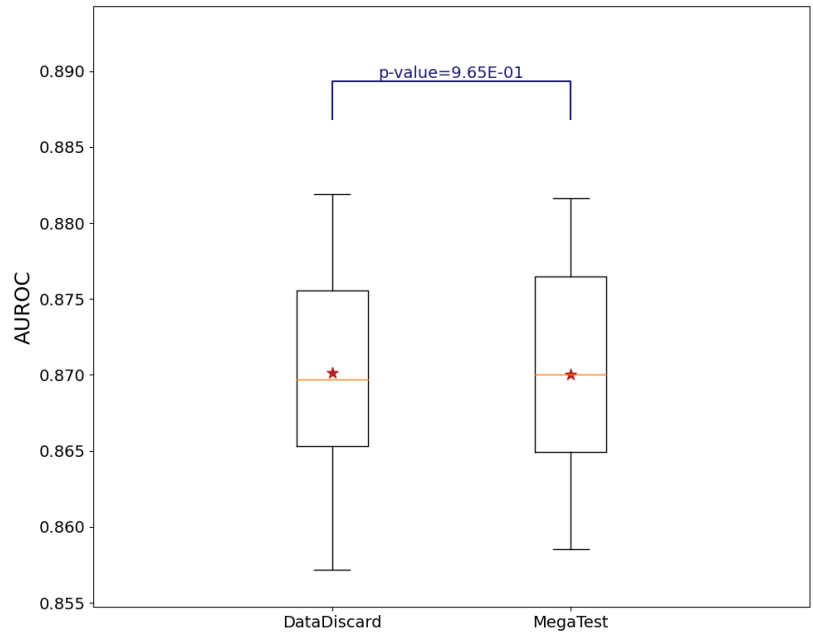

Figure 18. Testing with more examples (test size of 19,578 vs 185,863)

In the next step, we created a SQLite database and conducted feature engineering. The three versions of feature engineering were not statistically significant and had similar run times. Thus we chose version 2. This improved the mean AUROC from 0.870 to 0.962 which is exceptional and highlights the importance of transaction data. The run time after feature engineering was increased from 4 minutes to 7 minutes.

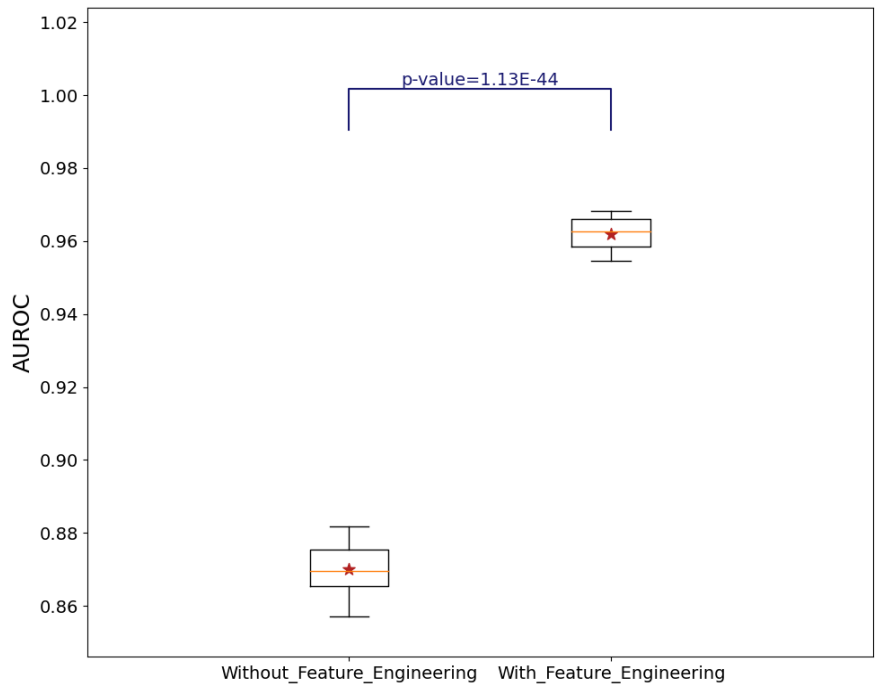

Figure 19. The effect of feature engineering

Switching back to a single data split (stratified 10%/90% test-to-dev) with 10-fold cross-validation for hyperparameter tuning enabled reporting more evaluation metrics and conducting XAI to derive feature importances. The model achieved an AUROC of 0.962, accuracy of 0.913, precision of 0.915, recall of 0.910, and F1 score of 0.913. We also derived the optimal hyperparameters ('n_estimators': 200, 'learning_rate': 0.2, 'num_leaves': 62, 'max_depth': 5, 'reg_lambda': 1, 'max_bin': 510, 'boosting_type': 'gbdt', 'objective': 'binary', 'metric': 'auc', 'is_unbalance': True) and achieved an impressive confusion matrix (Fig. 20).

Total number of wire transfers appeared as the most significant predictor (Fig. 21). Among the KYC features, age was the most significant one. Nevertheless, it should be highlighted that with one-hot encoding, the importance of occupation is split among multiple features.

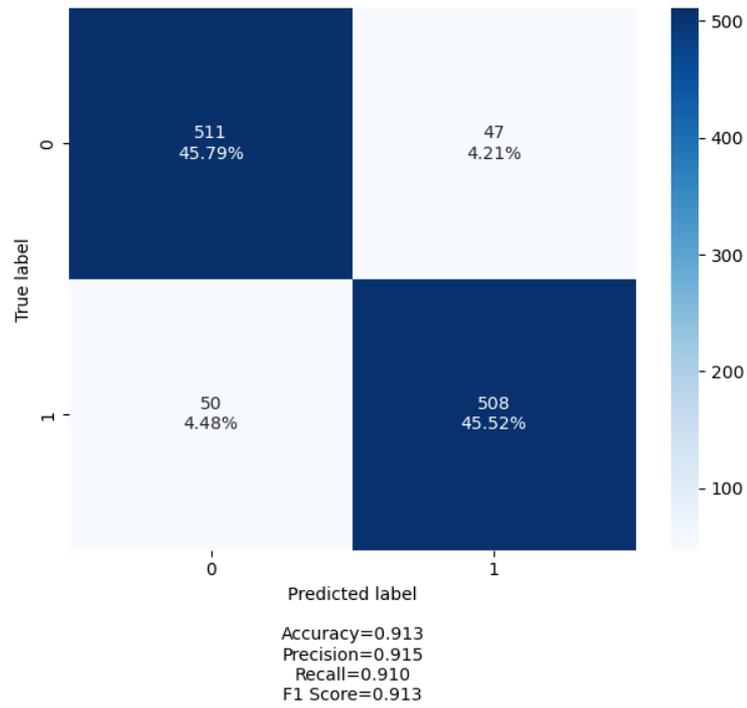

Figure 20. Confusion matrix of the final model

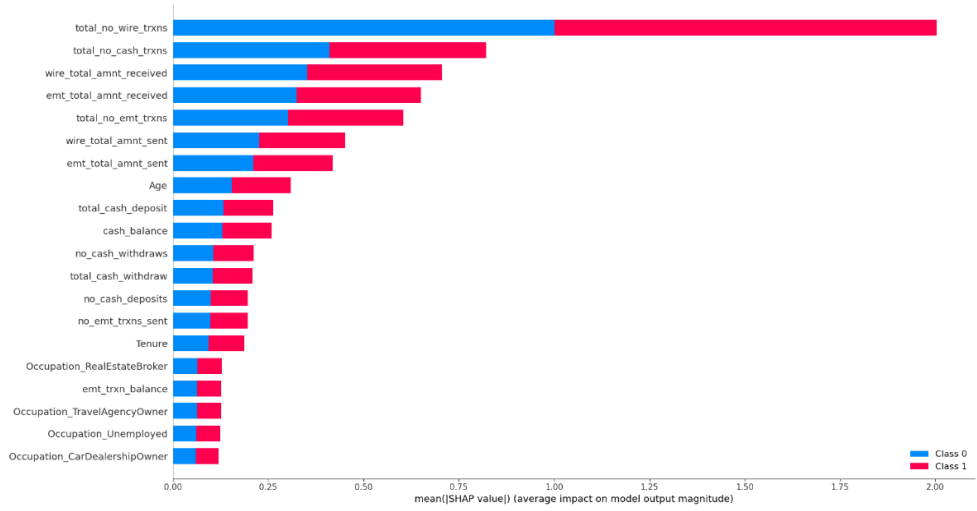

Figure 21. Feature importance for the final model based on Shaply values

## Conclusion

Our findings reveal that ML models excel in detecting high-risk clients for financial institutions, achieving remarkable accuracy and swift processing times. The use of SQL enhances data organization and streamlines the feature engineering process. By integrating ML models with AML SQL databases, we leverage the power of big data to make precise, real-time predictions on previously unseen data, continuously improving over time. This synergy not only optimizes the detection capabilities but also ensures that financial institutions can proactively respond to emerging threats with enhanced efficiency. The proposed pipeline was awarded second place in the competition.

## Acknowledgments

This research has been made possible by accessing the dataset curated by Scotiabank's data scientists. We appreciate the technical support of Duncan Halverson and Parsa Vafaie. Also, we appreciate the efforts of Kevin Yousie and others from the Institute for Management and Innovation, University of Toronto, who facilitated the competition.

## Declaration of Interest Statement

The authors declare no competing interests.

## References

[1] M. Jullum, A. Løland, R. B. Huseby, G. Ånonsen, and J. Lorentzen, "Detecting money laundering transactions with machine learning," *Journal of Money Laundering Control*, vol. 23, no. 1, pp. 173–186, Jan. 2020, doi: 10.1108/JMLC-07-2019-0055.

[2] I. Alarab, S. Prakoonwit, and M. I. Nacer, "Comparative Analysis Using Supervised Learning Methods for Anti-Money Laundering in Bitcoin," in *Proceedings of the 2020 5th International Conference on Machine Learning Technologies*, in ICMLT '20. New York, NY, USA: Association for Computing Machinery, 2020, pp. 11–17. doi: 10.1145/3409073.3409078.

[3] K. Sintayehu and H. Seid, "Developing Anti Money Laundering Identification using Machine Learning Techniques," *Irish Interdisciplinary Journal of Science & Research*, 2023, doi: 10.46759/iijsr.2023.7110.

[4] H. M. Sani, C. Lei, and D. Neagu, "Computational Complexity Analysis of Decision Tree Algorithms," pp. 191–197, 2018, doi: 10.1007/978-3-030-04191-5_17.

[5] K. Namdar, M. A. Haider, and F. Khalvati, "A Modified AUC for Training Convolutional Neural Networks: Taking Confidence Into Account," *Front Artif Intell*, vol. 4, p. 155, 2021, doi: 10.3389/frai.2021.582928.

[6] K. Namdar, M. W. Wagner, B. B. Ertl-Wagner, and F. Khalvati, "Open-radiomics: A Research Protocol to Make Radiomics-based Machine Learning Pipelines Reproducible." 2022.

[7] M. W. Wagner *et al.*, "Radiomics of pediatric low grade gliomas: toward a pretherapeutic differentiation of BRAF-mutated and BRAF-fused tumors," *American Journal of Neuroradiology*, 2021.

[8] T. Chen and C. Guestrin, "XGBoost: A Scalable Tree Boosting System," in *Proceedings of the 22nd ACM SIGKDD International Conference on Knowledge Discovery and Data Mining*, in KDD '16. New York, NY, USA: ACM, 2016, pp. 785–794. doi: 10.1145/2939672.2939785.

[9] L. Prokhorenkova, G. Gusev, A. Vorobev, A. V. Dorogush, and A. Gulin, "CatBoost: unbiased boosting with categorical features." 2017.

[10] G. Ke *et al.*, "LightGBM: A Highly Efficient Gradient Boosting Decision Tree," in *Advances in Neural Information Processing Systems*, I. Guyon, U. Von Luxburg, S. Bengio, H. Wallach, R. Fergus, S. Vishwanathan, and R. Garnett, Eds., Curran Associates, Inc., 2017. [Online]. Available: https://proceedings.neurips.cc/paper_files/paper/2017/file/6449f44a102fde848669bdd9eb6b76fa-Paper.pdf

[11] S. O. Arik and T. Pfister, "TabNet: Attentive Interpretable Tabular Learning." 2019.

[12] X. He, K. Zhao, and X. Chu, "AutoML: A Survey of the State-of-the-Art," 2019, doi: 10.1016/j.knosys.2020.106622.

[13] N. Erickson *et al.*, "AutoGluon-Tabular: Robust and Accurate AutoML for Structured Data." 2020.

[14] N. Chawla, K. Bowyer, L. Hall, and W. Kegelmeyer, "SMOTE: Synthetic Minority Over-sampling Technique," *ArXiv*, vol. abs/1106.1, 2002, doi: 10.1613/jair.953.

[15] M. Wagner *et al.*, "Dataset Size Sensitivity Analysis of Machine Learning Classifiers to Differentiate Molecular Markers of Pediatric Low-Grade Gliomas Based on MRI." Research Square, 2021. doi: 10.21203/rs.3.rs-883606/v1.